\documentclass[10pt,twocolumn,letterpaper]{article}
\usepackage{iccv}
\usepackage{times}
\usepackage{epsfig}
\usepackage{relsize}
\usepackage{subfigure}
\usepackage{graphicx}
\usepackage{amsmath}
\usepackage{amssymb}
\usepackage{multirow}
\usepackage{gensymb}
\usepackage{tabularx, booktabs}
\newcolumntype{Y}{>{\arraybackslash}X}
\usepackage{pifont}
\usepackage{caption}
\def\mbf#1{\mathbf{#1}}

\def\tbf#1{\textbf{#1}}

\newcommand{\minisection}[1]{\vspace{1mm}\noindent{\bf #1}.}

\DeclareGraphicsExtensions{.pdf,.jpg}

\usepackage[pagebackref=true,breaklinks=true,letterpaper=true,colorlinks,bookmarks=false]{hyperref}

\iccvfinalcopy
\begin{document}

\title{On Face Segmentation, Face Swapping, and Face Perception\vspace{-4mm}}

\author{Yuval Nirkin$^{1}$, Iacopo Masi$^2$, Anh Tu$\acute{\hat{\textnormal{a}}}$n Tr$\grave{\hat{\textnormal{a}}}$n$^2$, Tal Hassner$^{1,3}$, and G\'{e}rard Medioni$^2$\\
{\small $^{1}$ The Open University of Israel, Israel}\\
{\small $^{2}$ Institute for Robotics and Intelligent Systems, USC, CA, USA}\\
{\small $^{3}$ Information Sciences Institute, USC, CA, USA}
}

\makeatletter
\let\@oldmaketitle\@maketitle\renewcommand{\@maketitle}{\@oldmaketitle  \vspace{-0.5cm}\includegraphics[width=\linewidth,clip,trim = 0mm 0mm 0mm 0mm]{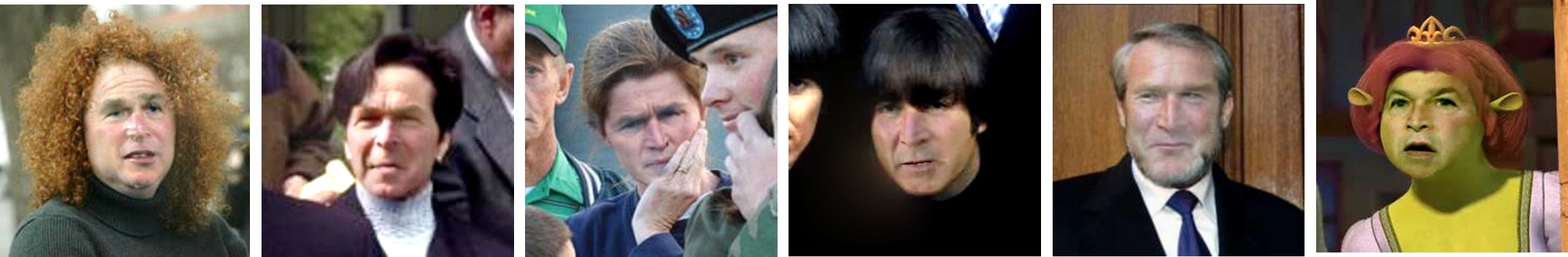}\captionof{figure}{{\em Inter-subject swapping.} LFW G.W. Bush photos swapped using our method onto very different subjects and images. Unlike previous work~\cite{bitouk2008face,kemelmacher2016transfiguring}, we do not select convenient targets for swapping. Is Bush hard to recognize? We offer quantitative evidence supporting Sinha and Poggio~\cite{sinha1996think} showing that faces and context are both crucial for recognition.}\label{fig:teaser}\bigskip}\makeatother

\maketitle

\begin{abstract}
We show that even when face images are unconstrained and arbitrarily paired, face swapping between them is actually quite simple. To this end, we make the following contributions. (a) Instead of tailoring systems for face segmentation, as others previously proposed, we show that a standard fully convolutional network (FCN) can achieve remarkably fast and accurate segmentations, provided that it is trained on a rich enough example set. For this purpose, we describe novel data collection and generation routines which provide challenging segmented face examples. (b) We use our segmentations to enable robust face swapping under unprecedented conditions. (c) Unlike previous work, our swapping is robust enough to allow for extensive quantitative tests. To this end, we use the Labeled Faces in the Wild (LFW) benchmark and measure the effect of intra- and inter-subject face swapping on recognition. We show that our intra-subject swapped faces remain as recognizable as their sources, testifying to the effectiveness of our method. In line with well known perceptual studies, we show that better face swapping produces less recognizable inter-subject results (see, e.g., Fig.~\ref{fig:teaser}). This is the first time this effect was quantitatively demonstrated for machine vision systems.
\end{abstract}

\section{Introduction}
Swapping faces means transferring a face from a {\em source} photo onto a face appearing in a {\em target} photo, attempting to generate realistic, unedited looking results. Although face swapping today is often associated with viral Internet memes~\cite{deza2015understanding,oikawa2016manifold}, it is actually far more important than this practice may suggest: Face swapping can also be used for preserving privacy~\cite{bitouk2008face,mosaddegh2014photorealistic,ross2011visual}, digital forensics~\cite{oikawa2016manifold} and as a potential face specific data augmentation method~\cite{masi16dowe} especially in applications where training data is scarce (e.g., facial emotion recognition~\cite{levi2015emotion}).

Going beyond particular applications, face swapping is also an excellent opportunity to develop and test essential face processing capabilities: When faces are swapped between arbitrarily selected, unconstrained images, there is no guarantee on the similarity of viewpoints, expressions, 3D face shapes, genders or any other attribute that makes swapping easy~\cite{kemelmacher2016transfiguring}. In such cases, swapping requires robust and effective methods for face alignment, segmentation, 3D shape estimation (though we will later challenge this assertion), expression estimation and more.  

We describe a face swapping method and test it in settings where no control is assumed over the images {\em or their pairings}. We evaluate our method using extensive quantitative tests at a scale never before attempted by other face swapping methods. These tests allow us to measure the effect face swapping has on machine face recognition, providing insights from the perspectives of both security applications and face perception.

Technically, we focus on face segmentation and the design of a face swapping pipeline. Our contributions include: 
\begin{itemize}
\item {\em Semi-supervised labeling of face segmentation.} We show how a rich image set with face segmentation labels can be generated with little effort by using motion cues and 3D data augmentation. The data we collect is used to train a standard FCN to segment faces, surpassing previous results in both accuracy and speed. 
\item {\em Face swapping pipeline.} We describe a pipeline for face swapping and show that our use of improved face segmentation and robust system components leads to high quality results even under challenging unconstrained conditions. 
\item {\em Quantitative tests.} Despite over a decade of work and contrary to other face processing tasks (e.g., recognition), face swapping methods were never quantitatively tested. We design two test protocols based on the LFW benchmark~\cite{LFWTech} to test how intra- and inter-subject face swapping affects face verification. 
\end{itemize}

Our qualitative results show that our swapped faces are as compelling as those produced by others, if not more. Our quantitative tests further show that our intra-subject face swapping has little effect on face verification accuracy: our swapping does not introduce artifacts or otherwise changes these images in ways which affect subject identities.

We report inter-subject results on randomly selected pairs. These tests require facial appearance to change, sometimes substantially, in order to naturally blend source faces into their new surroundings. We show that this changes them, making them less recognizable. Though this perceptual phenomenon was described over two decades ago by Sinha and Poggio~\cite{sinha1996think} in their well-known Clinton-–Gore illusion, we are unaware of previous quantitative reports on how this applies to machine face recognition. 

For code, deep models and more information, please see our project webpage.\footnote{\url{www.openu.ac.il/home/hassner/projects/faceswap}}

\section{Related Work}\label{sec:related}

\minisection{Face segmentation} To swap only faces, without their surrounding context or occlusions, we require per-pixel segmentation labels. Previous methods designed for this purpose include the work of~\cite{luo2012hierarchical} which segment individual facial regions (e.g., eyes, mouth) but not the entire face. An example based method was proposed in~\cite{smith2013exemplar}. More recently,~\cite{ghiasi2015using} segmented faces by alternating between segmentation and landmark localization using deformable part
models. They report state of the art performance on the Caltech Occluded Faces in the Wild (COFW) dataset~\cite{burgos2013robust}. 

Two recent methods proposed to segment faces using deep neural networks. In~\cite{liu2015multi} a network was trained to simultaneously segment multiple facial regions, including the entire face. This method was used in the face swapping method of~\cite{kemelmacher2016transfiguring}, but can be slow. The very recent method of~\cite{saito2016real} recently outperformed~\cite{ghiasi2015using} on COFW as well as reported real-time processing speeds by using a deconvolutional neural network. We use a FCN for segmentation, proposing novel training data collection and augmentation methods to obtain challenging training examples.

\minisection{Face swapping} Methods for swapping faces were proposed as far back as 2004 by~\cite{blanz2004exchanging} with fully automatic techniques described nearly a decade ago in~\cite{bitouk2008face}. These methods were originally offered in response to privacy preservation concerns: Face swapping can be used to obfuscate identities of subjects appearing in publicly available photos, as a substitute to face pixelation or blurring~\cite{bitouk2008face,blanz2004exchanging,de2010linear,lin2012face,mosaddegh2014photorealistic,ross2011visual}. Since then, however, many of their applications seem to come from recreation~\cite{kemelmacher2016transfiguring} or entertainment (e.g.,~\cite{alexander2009creating,wolf2010eye}).

Regardless of the application, previous face swapping systems often share several key aspects. First, some methods restrict the target photos used for transfer. Given an input source face, they search through large face albums to choose ones that are easy targets for face swapping~\cite{bitouk2008face,chen2013poseshop,kemelmacher2016transfiguring}. Such targets are those which share similar appearance properties with the source, including facial tone, pose, expression and more. Though our method can be applied in similar settings, our tests focus on more extreme conditions, where the source and target images are arbitrarily selected and can be (often are) substantially different. 

Second, most previous methods estimate the structure of the face. Some methods estimate 3D facial shapes~\cite{alexander2009creating,bitouk2008face,lin2012faceB,lin2012face}, by fitting 3D Morphable Face Models (3DMM). Others instead estimate dense 2D active appearance models~\cite{de2010linear,zhu2009unsupervised}. This is presumably done in order to correctly map textures across different individual facial shapes.

Finally, deep learning was used to transfer faces~\cite{korshunova2016fast}, as if they were styles  transfered between images. This method, however, requires the network to be trained for each source image and thus can be impractical in many applications.

\begin{figure*}[t]
\centering
\includegraphics[width=.97\textwidth]{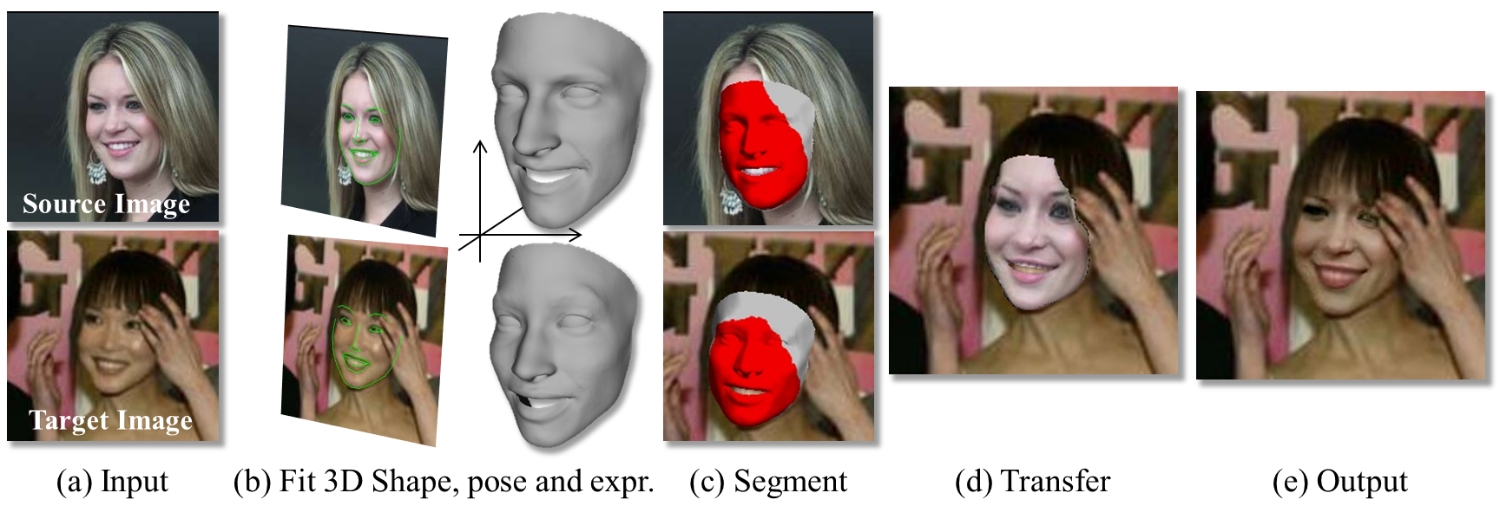}\vspace{-2mm}
\caption{
{\em Overview of our method.} (a) Source (top) and target (bottom) input images. (b) Detected facial landmarks used to establish 3D pose and facial expression for a 3D face shape (Sec.~\ref{sec:shapefitting}). We show the 3DMM regressed by~\cite{tran16_3dmm_cnn} but our tests demonstrate that a generic shape often works equally well. (c) Our face segmentation of Sec.~\ref{sec:segmentation} (red) overlaid on the projected 3D face (gray). (d) Source transfered onto target without blending, and the final results (e) after blending (Sec.~\ref{sec:swapping}).\vspace{-4mm} }\label{fig:system}
\end{figure*}

\section{Swapping faces in unconstrained images}\label{sec:method}

Fig.~\ref{fig:system} summarizes our face swapping method. When swapping a face from a source image, $\mathbf{I}_S$, to a target image, $\mathbf{I}_T$, we treat both images the same, apart from the final stage (Fig.~\ref{fig:system}(d)). Our method first localizes 2D facial landmarks in each image (Fig.~\ref{fig:system}(b)). We use an off-the-shelf detector for this purpose~\cite{kazemi2014one}. Using these landmarks, we compute 3D pose (viewpoint) and modify the 3D shape to account for expression. These steps are discussed in Sec.~\ref{sec:shapefitting}.

We then segment the faces from their backgrounds and occlusions (Fig.~\ref{fig:system}(c)) using a FCN trained to predict per-pixel face visibility (Sec.~\ref{sec:segmentation}). We describe how we generate rich labeled data to effectively train our FCN. Finally, the source is efficiently warped onto the target using the two, aligned 3D face shapes as proxies, and blended onto the target image (Sec.~\ref{sec:swapping}). 

\subsection{Fitting 3D face shapes}\label{sec:shapefitting}
To enrich our set of examples for training the segmentation network (Sec.~\ref{sec:segmentation}) we explicitly model 3D face shapes. These 3D shapes are also used as proxies to transfer textures from one face onto another, when swapping faces (Sec.~\ref{sec:swapping}). We experimented with two alternative methods of obtaining these 3D shapes. 

The first, inspired by~\cite{hassner2015effective} uses a generic 3D face, making no attempt to fit its shape to the face in the image aside from pose (viewpoint) alignment. We, however, also estimate facial expressions and modify the 3D face accordingly. 

A second approach uses the recent state of the art, deep method for single image 3D face reconstruction~\cite{tran16_3dmm_cnn}. It was shown to work well on unconstrained photos such as those considered here. To our knowledge, this is the only method quantitatively shown to produce invariant, discriminative and accurate 3D shape estimations. The code they released regresses 3D Morphable face Models (3DMM) in neutral pose and expression. We extend it by aligning 3D shapes with input photos and modifying the 3D faces to account for facial expressions. 

\minisection{3D shape representation and estimation} Whether generic or regressed, we use the popular Basel Face Model (BFM)~\cite{paysan09basel} to represent faces and the 3DDFA Morphable Model~\cite{Zhu2016Face} for expressions. These are both publicly available 3DMM representations. More specifically, a 3-D face shape~$\mathbf{V}\subset\mathbb{R}^3$ is modeled by combining the following independent generative models:
\begin{equation}
\mbf{V} = \widehat{\mbf{v}} + \mbf{W}_S~\boldsymbol{\alpha} + \mbf{W}_E~\boldsymbol{\gamma}.\label{eq:3DMM}
\end{equation}
Here, vector $\widehat{\mbf{v}}$ is the mean face shape, computed over aligned facial 3D scans in the Basel Faces collection and represented by the concatenated 3D coordinates of their 3D points. When using a generic face shape, we use this average face. Matrices $\mbf{W}_S$ (shape) and $\mbf{W}_E$ (expression) are principle components obtained from the 3D face scans. Finally, $\boldsymbol{\alpha}$ is a subject-specific 99D parameter vector estimated separately for each image and $\boldsymbol{\gamma}$ is a 29D parameter vector for expressions. To fit 3D shapes and expressions to an input image, we estimate these parameters along with camera matrices.

To estimate per-subject 3D face shapes, we regress $\boldsymbol{\alpha}$ using the deep network of~\cite{tran16_3dmm_cnn}. They jointly estimate 198D parameters for face shape and texture. Dropping the texture components, we obtain $\boldsymbol{\alpha}$ and back-project the regressed face by $\widehat{\mbf{v}} + \mbf{W}_S~\boldsymbol{\alpha}$, to get the estimated shape in 3D space.

\minisection{Pose and expression fitting}
Given a 3D face shape (generic or regressed) we recover its pose and adjust its expression to match the face in the input image. We use the detected facial landmarks, $\mbf{p}$ = $\{\mbf{p}_i\} \subset \mathbb{R}^{2}$, for both purposes. Specifically, we begin by solving for the pose, ignoring expression. We approximate the positions in 3D of the detected 2D facial landmarks $\mbf{\tilde{V}}$ = $\{\mbf{\tilde{V_i}}\}$  by:
\begin{equation}
\mbf{\tilde{V}} \approx f(\widehat{\mbf{v}}) + f(\mbf{W}_S)~\boldsymbol{\alpha},\label{eq:LM_shape}
\end{equation}
where $f(\cdot)$ is a function selecting the landmark vertices on the 3D model. The vertices of all BFM faces are registered so that the same vertex index corresponds to the same facial feature in all faces. Hence, $f$ need only be manually specified once, at preprocessing. From $f$ we get 2D-3D correspondences, $\mbf{p}_i \leftrightarrow \mbf{\tilde{V}}_i$, between detected facial features and their corresponding points on the 3D shape. Similarly to~\cite{hassner2013viewing}, we use these correspondences to estimate 3D pose, computing 3D face rotation, $\mbf{R} \in \mathbb{R}^{3}$, and translation vector $\mbf{t} \in \mathbb{R}^{3}$ using the EPnP solver~\cite{lepetit2009epnp}.

Following pose estimation, we estimate the expression parameters in vector $\boldsymbol{\gamma}$ by formulating expression estimation as a bounded linear problem:

{\setlength{\abovedisplayskip}{0pt}
\begin{flalign}
~&\delta_R \Big( P(\mbf{R}, \mbf{t}) \big(f(\widehat{\mbf{v}}) + f(\mbf{W}_S) ~\boldsymbol{\alpha} +  f(\mbf{W}_E) ~\boldsymbol{\gamma} \big) \Big) = \delta_R(\mbf{p}),\nonumber\\
~&\text{with} \left|\boldsymbol{\gamma}_j \right| \leq 3~{\sigma}_j  \quad \quad  \forall~j=\{1\ldots29\}\label{eq:Expression}
\end{flalign}
}where $\delta_R(\cdot)$ is a visibility check that removes occluded points given the head rotation $\mbf{R}$; $P(\mbf{R}, \mbf{t})$ is the projection matrix, given the extrinsic parameters ($\mbf{R}$,$\mbf{t}$); and ${\sigma}_j$ is the standard deviation of the $j$-th expression component in~$\boldsymbol{\gamma}$. This problem can be solved using any constrained linear least-squares solver.
\subsection{Deep face segmentation}\label{sec:segmentation}
Our method uses a FCN to segment the visible parts of faces from their context and occlusions. Other methods previously tailored novel network architectures for this task (e.g.,~\cite{saito2016real}). We show that excellent segmentation results can be obtained with a standard FCN, provided that it is trained on plenty of rich and varied examples. 

Obtaining enough diverse images with ground truth segmentation labels can be hard:~\cite{saito2016real}, for example, used manually segmented LFW faces and the semi-automatic segmentations of~\cite{cao_facew_TVCG}. These labels were costly to produce and limited in their variability and number. We obtain numerous training examples with little manual effort and show that a {\em standard FCN} trained on these examples outperforms state of the art face segmentation results.

\minisection{FCN architecture} 
We used the FCN-8s-VGG architecture, fine-tuned for segmentation on PASCAL by~\cite{long_shelhamer_fcn}.
Following~\cite{long_shelhamer_fcn}, we fuse information at different locations from layers with different strides. We refer to~\cite{long_shelhamer_fcn} for more details on this. 

\minisection{Semi-supervised training data collection} We produce large quantities of segmentation labeled face images by using {\em motion cues} in unconstrained face videos. To this end, we process videos from the recent IARPA Janus CS2 dataset~\cite{Klare_2015_CVPR}. These videos portray faces of different poses, ethnicities and ages, viewed under widely varying conditions. We used 1,275 videos of subjects not included in LFW, of the 2,042 CS2 videos (309 subjects out of 500).

Given a video, we produce a rough, initial segmentation using a method based on~\cite{Grundmann_VideoSegm_cvpr10}. Specifically, we keep a hierarchy of regions with stable region boundaries computed with dense optical flow. Though these regions may be over- or under-segmented, they are computed with temporal coherence and so these segments are consistent across frames.

\begin{figure}[tb]
\centering
\includegraphics[width=.95\linewidth,clip,trim = 0mm 0mm 0mm 0mm]{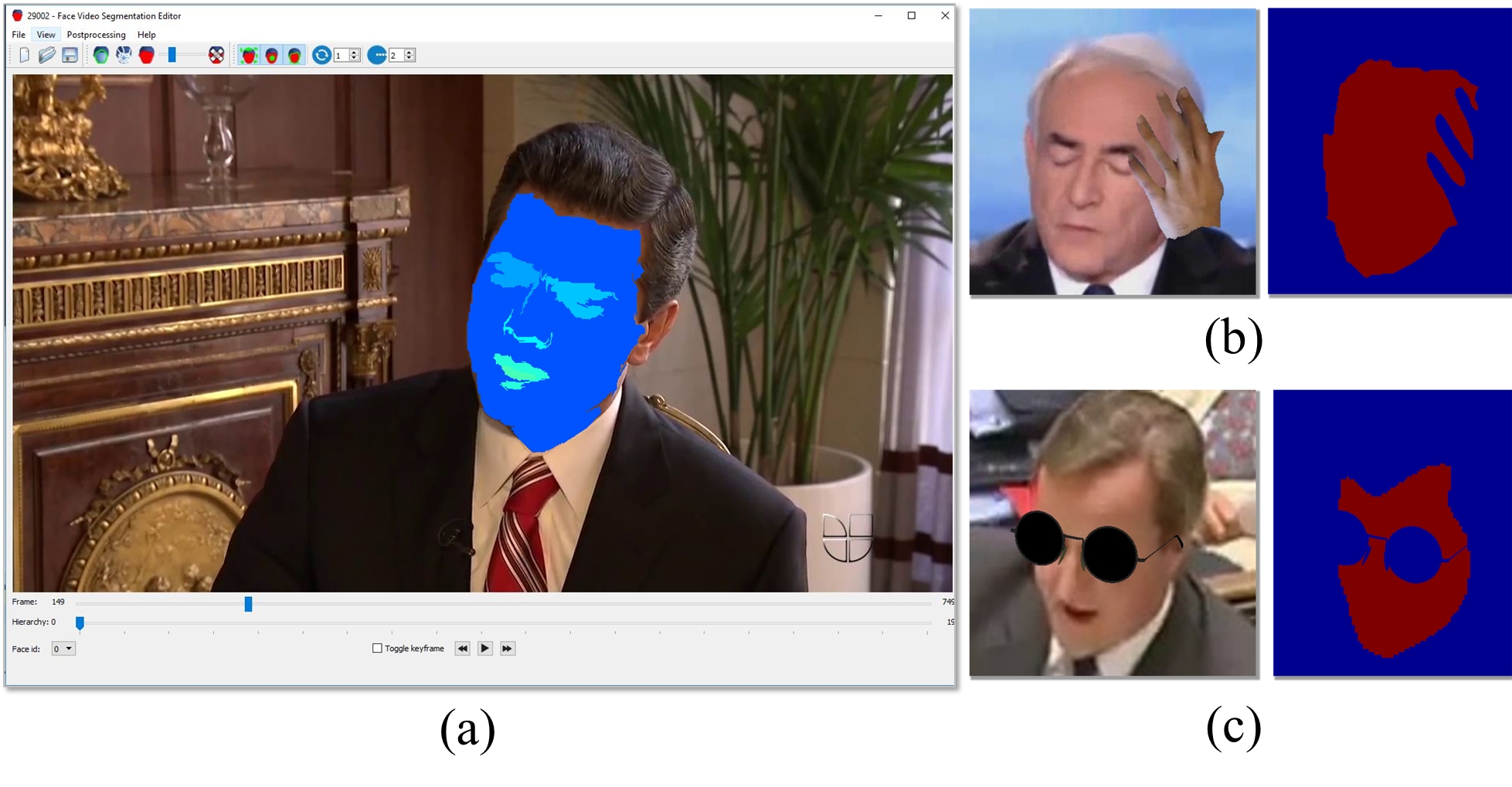}\vspace{-2mm}
\caption{
(a) Interface used for semi-supervised labeling. (b-c) Augmented examples and segmentation labels for occlusions due to (b) hands and (c) synthetic sunglasses. \vspace{-4mm}}\label{fig:interface}
\end{figure}

We use the method of~\cite{kazemi2014one} to detect faces and facial landmarks in each of the frames. Facial landmarks were then used to extract the face contour and extend it to include the forehead. All the segmented regions generated above, that did not overlap with a face contour are then discarded.  All intersecting segmented regions are further processed using a simple interface which allows browsing the entire video, selecting the partial segments of~\cite{Grundmann_VideoSegm_cvpr10} and adding or removing them from the face segmentation using simple mouse clicks. Fig.~\ref{fig:interface}(a) shows the interface used in the semi-supervised labeling. A selected frame is typically processed in about five seconds. In total, we used this method to produce~9,818 segmented faces, choosing anywhere between one to five frames from each video, in a little over a day of work. 

\minisection{Occlusion augmentation}
This collection is further enriched by adding synthetic occlusions. To this end, we explicitly use 3D information estimated for our example faces. Specifically, we estimate 3D face shape for our segmented faces, using the method described in Sec.~\ref{sec:shapefitting}. We then use computer graphic (CG) 3D models of various objects (e.g., sunglasses) to modify the faces. We project these CG models onto the image and record their image locations as synthetic occlusions. Each CG object added 9,500 face examples. The detector used in our system~\cite{kazemi2014one} failed to accurately localize facial features on the remaining 318 faces, and so this augmentation was not applied to them. 

Finally, an additional source of synthetic occlusions was supplied  following~\cite{saito2016real} by overlaying hand images at various positions on our example images. Hand images were taken from the egohands dataset of~\cite{Bambach_2015_ICCV}. Fig~\ref{fig:interface}(b) shows a synthetic hand augmentation and Fig~\ref{fig:interface}(c) a sunglasses augmentation, along with their resulting segmentation labels.

\begin{figure*}[t]
\centering
\includegraphics[width=1.0\linewidth]{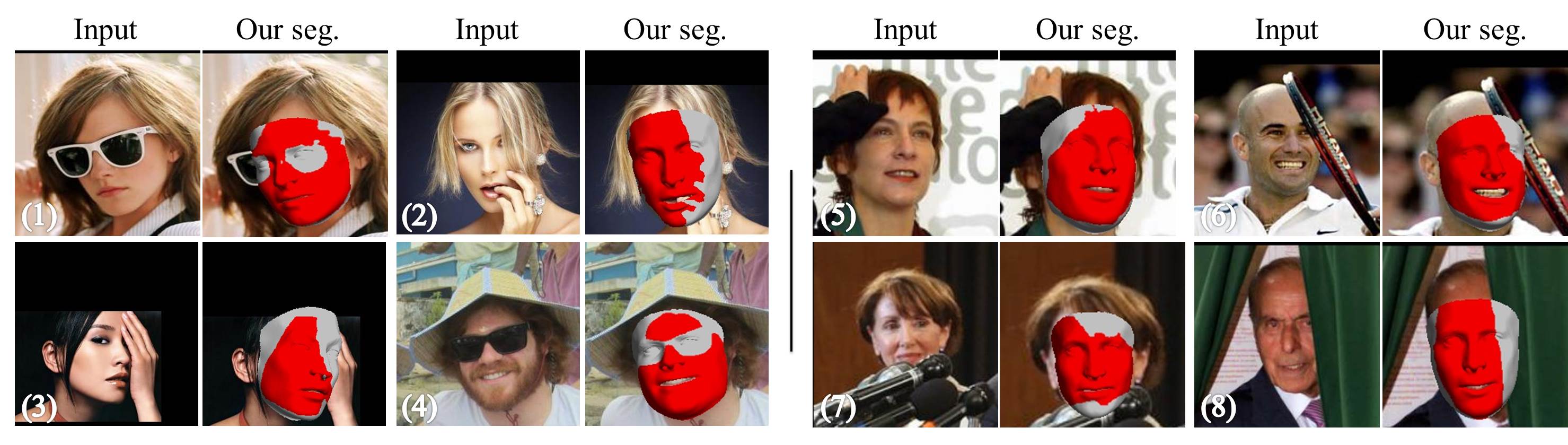}\vspace{-2mm}
\caption{
{\em Qualitative segmention results from the COFW (1-4) and LFW (5-8) data sets.}\vspace{-4mm}}\label{fig:qual_seg} \end{figure*}

\subsection{Face swapping and blending}\label{sec:swapping}
Face swapping from a source $\mbf{I}_S$ to target $\mbf{I}_T$ proceeds as follows. The 3D shape associated with the source, $\mathbf{V}_S$, is projected down onto $\mathbf{I}_S$ using its estimated pose, $P(\mathbf{R}_S,\mathbf{t}_S)$ (Sec.~\ref{sec:shapefitting}). We then sample the source image using bilinear interpolation, to assign 3D vertices projected onto the segmented face (Sec.~\ref{sec:segmentation}) with intensities sampled from the image at their projected coordinates.

The shapes for both source and target, $\mathbf{V}_S$ and $\mathbf{V}_T$ correspond in the indices of their vertices. We can therefore directly transfer these sampled intensities from all vertices $\mathbf{v}_i\in\mathbf{V}_S$ to $\mathbf{v}_i\in \mathbf{V}_T$. This provides texture for the vertices corresponding to visible regions in $\mathbf{I}_S$ on the target 3D shape. We now render $\mathbf{V}_T$ onto $\mathbf{I}_T$, using the estimated target pose $(\mathbf{R}_T,\mathbf{t}_T)$, masking the rendered intensities using the target face segmentation (see Fig.~\ref{fig:system}(d)). Finally, the rendered, source face is blended-in with the target context using an off the shelf method~\cite{Perez03a}.

\section{Experiments}\label{sec:experiments}
We performed comprehensive experiments in order to test our method, both qualitatively and quantitatively. Our face swapping method was implemented using MatConvNet~\cite{vedaldi15matconvnet} for segmentation, DLIB~\cite{dlib09} for facial landmark detection and OpenCV~\cite{itseez2015opencv} for all other image processing tasks. Runtimes were all measured on an Intel Core i7 4820K computer with 32GB DDR4 RAM and an NVIDIA GeForce Titan X. Using the GPU, our system swaps faces at 1.3 fps. On the CPU, this is slightly slower, performing at 0.8 fps. We emphasize again that unlike previous work our implementation will be public.

\begin{table}[t]
\centering
\resizebox{1.0\linewidth}{!}{
\begin{tabular}{l c@{~}c@{~}c@{~}c}
\toprule
Method & mean IOU  &	Global &	ave(face) & FPS  \\
\cmidrule(r){1-1} \cmidrule(l){2-5}
Struct. Forest~\cite{jia2014structured}$^*$ & -- & 83.9 & 88.6 & -- \\
RPP~\cite{Yang_tip_land} & 72.4 & -- & --  & 0.03 \\
SAPM~\cite{ghiasi2015using} & 83.5 & 88.6 & 87.1  & -- \\
Liu \emph{et al.}~\cite{liu2015multi} & 72.9 & 79.8 & 89.9 & 0.29\\Saito \emph{et al.}~\cite{saito2016real}$^*$ \emph{\small{+GraphCut}} & {\bf 83.9} & 88.7 & 92.7 & 43.2 \\
\cmidrule(r){1-1}
\tbf{Us}  & 83.7 & {\bf 88.8} & {\bf 94.1}  &   {\bf 48.6}\\\bottomrule
\end{tabular}
}
\caption{{\em COFW~\cite{burgos2013robust} segmentation results.}~*~\cite{saito2016real,Yang_tip_land} reported results on unspecified subsets of the test set.\vspace{-4mm}}
\label{tab:seg}
\end{table}

\begin{figure*}[t]
\centering
\includegraphics[width=.99\textwidth]{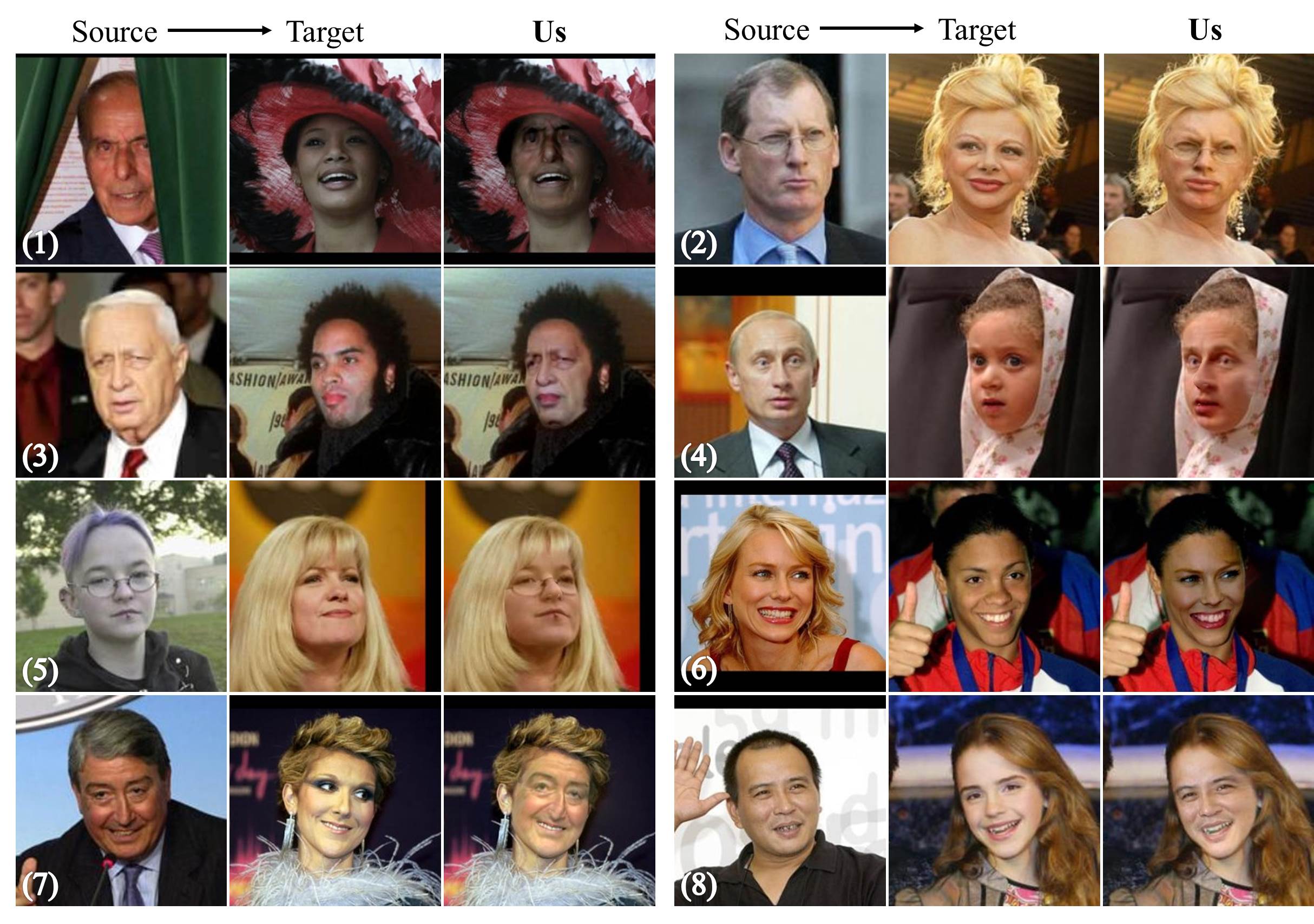}\vspace{-2mm}
\caption{
{\em Qualitative LFW inter-subject face swapping results.} Examples were selected to represent extremely different {\em poses} (4,7,8), {\em genders} (1,2,7,8), {\em expressions} (1,7), {\em ethnicities} (1,3,6,8), {\em ages} (3-8) and {\em occlusions} (1,5).\vspace{-3mm}}\label{fig:qual_inter}
\end{figure*}

\begin{figure}[tb]
\centering
\includegraphics[width=0.98\linewidth,clip,trim = 0mm 0mm 0mm 0mm]{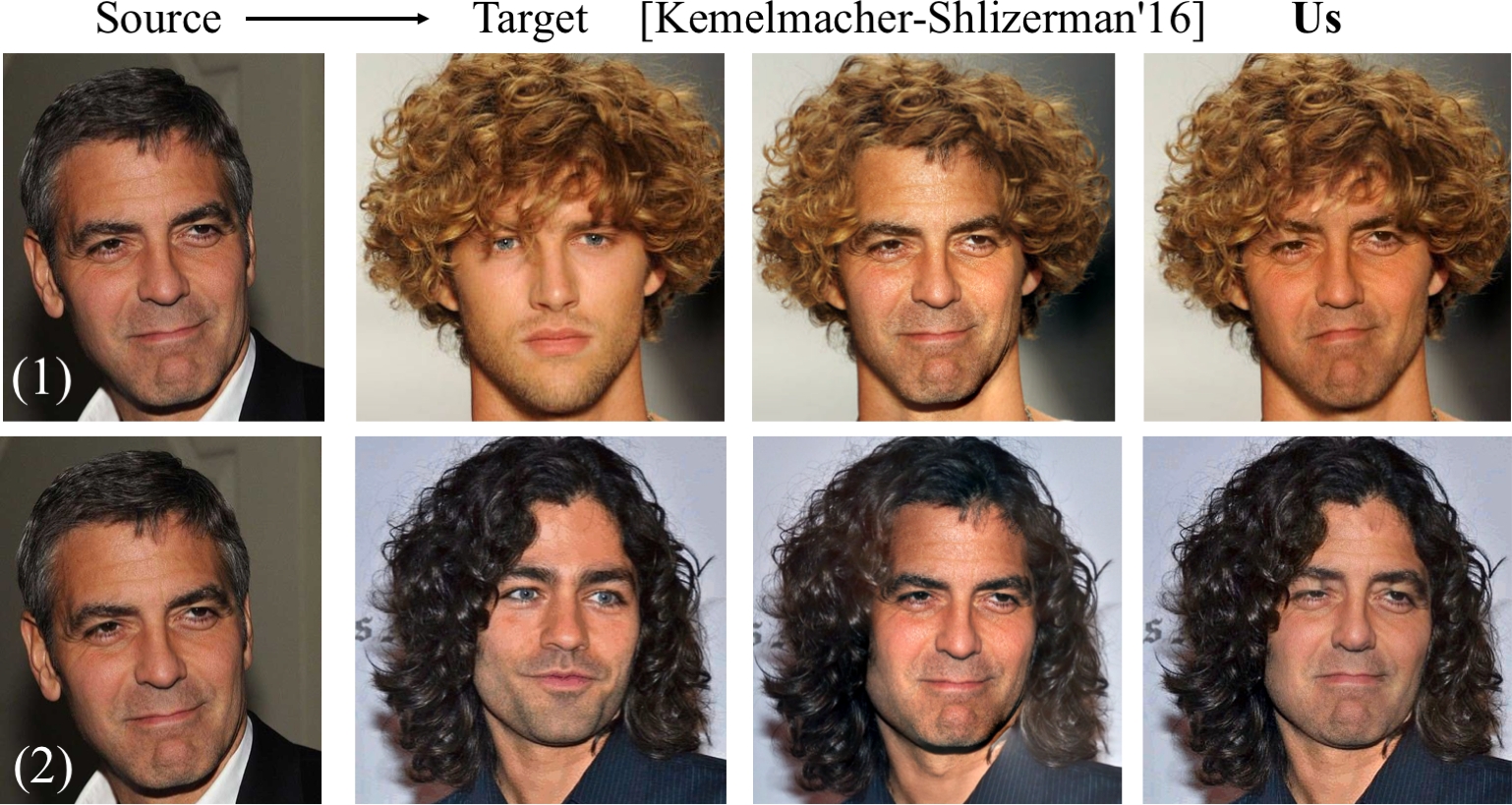}\vspace{-2mm}
\caption{
{\em Comparison with~\cite{kemelmacher2016transfiguring} using the same pairs.}\vspace{-3mm}}\label{fig:qual_ira}
\end{figure}

\begin{figure*}[t]
\centering
\includegraphics[width=.99\linewidth,clip, trim=1mm 1mm 1mm 1mm]{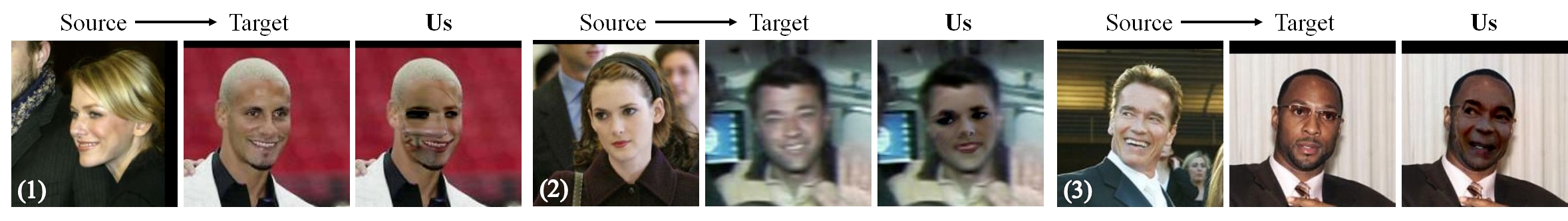}\vspace{-2mm}
\caption{
{\em Face swapping failures.} (1) The most common reason for failure is facial landmark localization errors, leading to misaligned shapes or poor expression estimation. Other, less frequent reasons include (2) substantially different image resolutions (3) failures in blending very different facial hues.\vspace{-4mm}}\label{fig:qual_fail}
\end{figure*}

\subsection{Face segmentation results}\label{sec:segmentationtests}
Qualitative face segmentation results are provided in Fig.~\ref{fig:system} and~\ref{fig:qual_seg}, visualized following~\cite{saito2016real} to show segmented regions (red) overlaying the aligned 3D face shapes, projected onto the faces (gray). 

We provide also quantitative tests, comparing the accuracy of our segmentations to existing methods. We follow the evaluation procedure described by~\cite{ghiasi2015using}, testing the 507 face photos in the COFW dataset~\cite{burgos2013robust}. Previous methods included the regional predictive power (RPP) estimation~\cite{Yang_tip_land}, Structured Forest~\cite{jia2014structured}, segmentation-aware part model (SAPM)~\cite{ghiasi2015using}, the deep method of~\cite{liu2015multi}, and~\cite{saito2016real}. Note that Structured Forest~\cite{jia2014structured} and~\cite{saito2016real} used respectively 300 and 437 images for testing, without reporting which images were used. Result for~\cite{liu2015multi} was computed by us, using their code, out of the box, but optimizing for the segmentation threshold which provided the best accuracy.

Accuracy is measured using the standard {\em intersection over union} (IOU) metric, comparing predicted segmentations with manually annotated ground truth masks from~\cite{jia2014structured}, as well as two metrics from~\cite{jia2014structured}: {\em global} -- overall percent of correctly labeled pixels -- and {\em ave(face)}, the average face pixel recall. Tab.~\ref{tab:seg} reports these results along with run times. Our method is the fastest yet achieves comparable result with the state of the art. Note that we use the same GPU model as~\cite{saito2016real} and measure run time for~\cite{liu2015multi} ourselves. Other run times were reported in previous work. 

\subsection{Qualitative face-swapping results}\label{sec:qualitative}
We provide face swapping examples produced on unconstrained LFW images~\cite{LFWTech} using randomly selected targets in Fig.~\ref{fig:teaser},~\ref{fig:system}, and~\ref{fig:qual_inter}. We chose these examples to demonstrate a variety of challenging settings. In particular, these results used source and target faces of widely different poses, occlusions and facial expressions. To our knowledge, previous work never showed results for such challenging settings.

In addition, Fig.~\ref{fig:qual_ira} shows a qualitative comparison with the very recent method~\cite{kemelmacher2016transfiguring} using the same source-target pairs. We note that~\cite{kemelmacher2016transfiguring} used the segmentation of~\cite{liu2015multi} which we show in Sec.~\ref{sec:segmentationtests} to perform worst than our own. This is qualitatively evident in Fig.~\ref{fig:qual_ira} by the face hairlines. Finally, Fig.~\ref{fig:qual_fail} describes some typical failure cases and their causes.

\subsection{Quantitative tests}\label{sec:quantitative}
Similarly to previous work, we offer qualitative results to visualize the realism and quality of our swapped faces (Sec.~\ref{sec:qualitative}). Unlike previous work, however, we also offer extensive quantitative tests designed to measure the effect of swapping on the perceived identity of swapped faces. To this end we propose two test protocols, motivated by the following assumptions.\\

\noindent{\em Assumption 1.} Swapping faces between images of different subjects (i.e., {\em inter-subject swapping}) changes the {\em context} of the face (e.g., hair, skin tone, head shape). Effective swapping must therefore modify source faces, sometimes substantially, to blend them naturally into their new contexts thereby producing faces that look less like the source subjects. \\

\noindent{\em Assumption 2.} If a face is swapped from source to target photos of the same person ({\em intra-subject swapping}), the output of an effective swapping method should easily be recognizable as the person in the source photo as the two photos share the same context. \\

The first assumption is based on well-known trait of human visual perception: Face recognition requires both internal and external cues (faces and their context) to recognize faces. This idea was claimed by the seminal work of~\cite{sinha1996think} and extensively studied in the context of biological visual systems (e.g.,~\cite{axelrod2010external}). To our knowledge, it was never explored for machine recognition systems and never quantitatively. The robustness of our method allows us to do just that. 

The second assumption is intended to verify that when the context remains the same (the same subject) swapping does not change facial appearances in a way which makes faces less recognizable. This ensures that the swapping method does not introduce artifacts or unnecessary changes to facial appearances. 

To test these assumptions, we produce modified (face swapped) versions of the LFW benchmark~\cite{LFWTech}. We estimate how recognizable faces appear after swapping by using a publicly available, state of the art face recognition system in lieu of a large scale human study. Though the recognition abilities of humans and machines may be different, modern systems already claim human or even super-human accuracy~\cite{lu2014surpassing}. We therefore see the use of a state of the art machine system as an adequate surrogate to human studies which often involve problems of their own~\cite{learned2016labeled}.

\minisection{Face verification}
We use the ResFace101~\cite{masi16dowe} face recognition system to test if faces remain recognizable after swapping. ResFace101 obtained near perfect verification results on LFW, yet it was not optimized for that benchmark and tested also on IJB-A~\cite{Klare_2015_CVPR}. Moreover, it was trained on synthetic face views not unlike the ones produced by face swapping. For these reasons, we expect ResFace101 to be well suited for our purposes. Recognition is measured by 100\%-EER (Equal Error Rate), accuracy (Acc.), and normalized Area Under the Curve (nAUC). Finally, we provide ROC curves for all our tests.

\minisection{Inter-subject swapping verification protocols}
We begin by measuring the effect of inter-subject face swapping on face verification accuracy. To this end, we process all faces in the LFW benchmark, swapping them onto photos of {\em other, randomly selected subjects}. We make no effort to verify the quality of the swapped results and if swapping failed (e.g., Fig.~\ref{fig:qual_fail}), we treat the result as any other image. 

We use the original LFW test protocol with its same/not-same subject pairs. Our images, however, present the original faces with possibly very different contexts. Specifically, let $(\mbf{I}_i^1,\mbf{I}_i^2)$ be the $i$-th LFW test image pair. We produce $\widehat{\mbf{I}}_i^1$, the swapped version of $\mbf{I}_i^1$, by randomly picking another LFW subject and image from that subject as a target, taking $\mbf{I}_i^1$ as the source. We then do the same for $\mbf{I}_i^2$ to obtain $\widehat{\mbf{I}}_i^2$.

Ostensibly, we can now measure recognition accuracy using the new pairs: $(\widehat{\mbf{I}}_i^1, \widehat{\mbf{I}}_i^2)$. Matching pairs of swapped images, however, can obscure changes to both images which make the source faces equally unrecognizable: Such tests only reflect the similarity of swapped images to each other, not to their sources. We therefore test verification on benchmark pairs comparing original vs. swapped images. This is done twice, once on pairs $(\widehat{\mbf{I}}_i^1, {\mbf{I}}_i^2)$, the other on pairs $({\mbf{I}}_i^1, \widehat{\mbf{I}}_i^2)$. We then report the average results for both trials. We refer to these tests as {\em face preserving} tests, as swapping preserves the face, transferring it onto a new context.

We also performed {\em context preserving} tests. These tests use benchmark image pairs as {\em targets} and not sources as before. Thus the faces in these images preserve the context of the original LFW images, rather than the faces. By doing so, we can measure the effect of context on recognition. This test setup is reminiscent of the {\em inverse mask} tests performed by~\cite{kumar2009attribute}. Their tests were designed to measure how well humans recognize LFW faces if the face was cropped out without being replaced, and showed this led to a drop in recognition. Unlike their tests which used black boxes to occlude faces, our images contain faces of other subjects swapped in place of the original faces, and so our images are more challenging than theirs.

\begin{figure}[t]
\centering
\includegraphics[width=.91\linewidth]{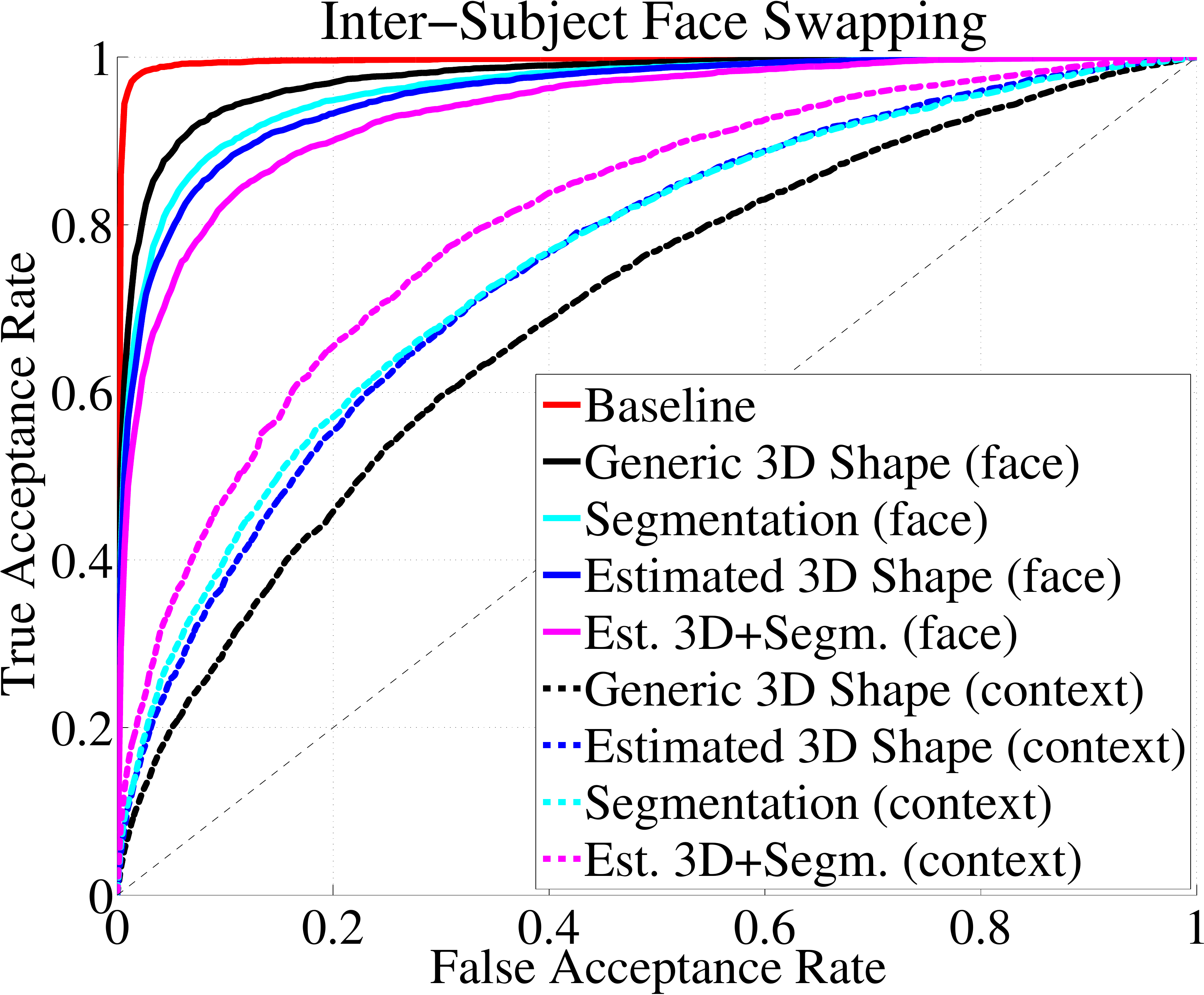}\vspace{-2mm}
\caption{
{\em Inter-subject swapping ROC curves.} Ablation study for the two experiments. Baseline shown in red.\vspace{-4mm}}\label{fig:rocs}
\end{figure}

\minisection{Inter-subject swapping results}
We provide verification results for both face preserving and context preserving inter-subject face swapping in Tab.~\ref{tab:inter} and ROC curves for the various tests in Fig.~\ref{fig:rocs}. Our results include ablation studies, showing accuracy with a generic face and no segmentation ({\em Generic}), with an estimated 3D face shape (Sec.~\ref{sec:shapefitting}) and no segmentation ({\em Est. 3D}), with a generic face and segmentation ({\em Seg.}) and with an estimated 3D shape and face segmentation ({\em Est. 3D+Seg.}).

The face preserving results in Tab.~\ref{tab:inter} (bottom) are consistent with our {\em Assumption~1}: The more the source face is modified, by estimating 3D shape and better segmenting the face, the less it is recognizable as the original subject and the lower the verification results. Using a simple generic shape and no segmentation provides $\sim$8\% {\em better} accuracy than using our the entire pipeline. Importantly, just by estimating 3D face shapes, accuracy drops by $\sim$3.5\% compared to using a simple generic face shape.

Unsurprisingly, the context preserving results in Tab.~\ref{tab:inter} (top) are substantially lower than the face preserving tests. Unlike the face preserving tests, however, the harder we work to blend the randomly selected source faces into their contexts, the better recognition becomes. By better blending the sources into the context, more of the context is retained and the easier it is to verify the two images based on their context without the face itself misleading the match.

\minisection{Intra-subject swapping verification protocols and results}
To test our second assumption, we again process the LFW benchmark, this time swapping faces between different images of the {\em same subjects} ({\em intra-subject} face swapping). Of course, all {\em same} labeled test pairs, by definition, belong to subjects that have at least two images, and so this did not affect these pairs. {\em Not-same} pairs, however, sometimes include images from subjects which have only a single image. To address this, we replaced them with others for which more than one photo exists. 

\begin{figure}[tb]
\centering
\includegraphics[width=0.98\linewidth,clip,trim = 0mm 0mm 0mm 0mm]{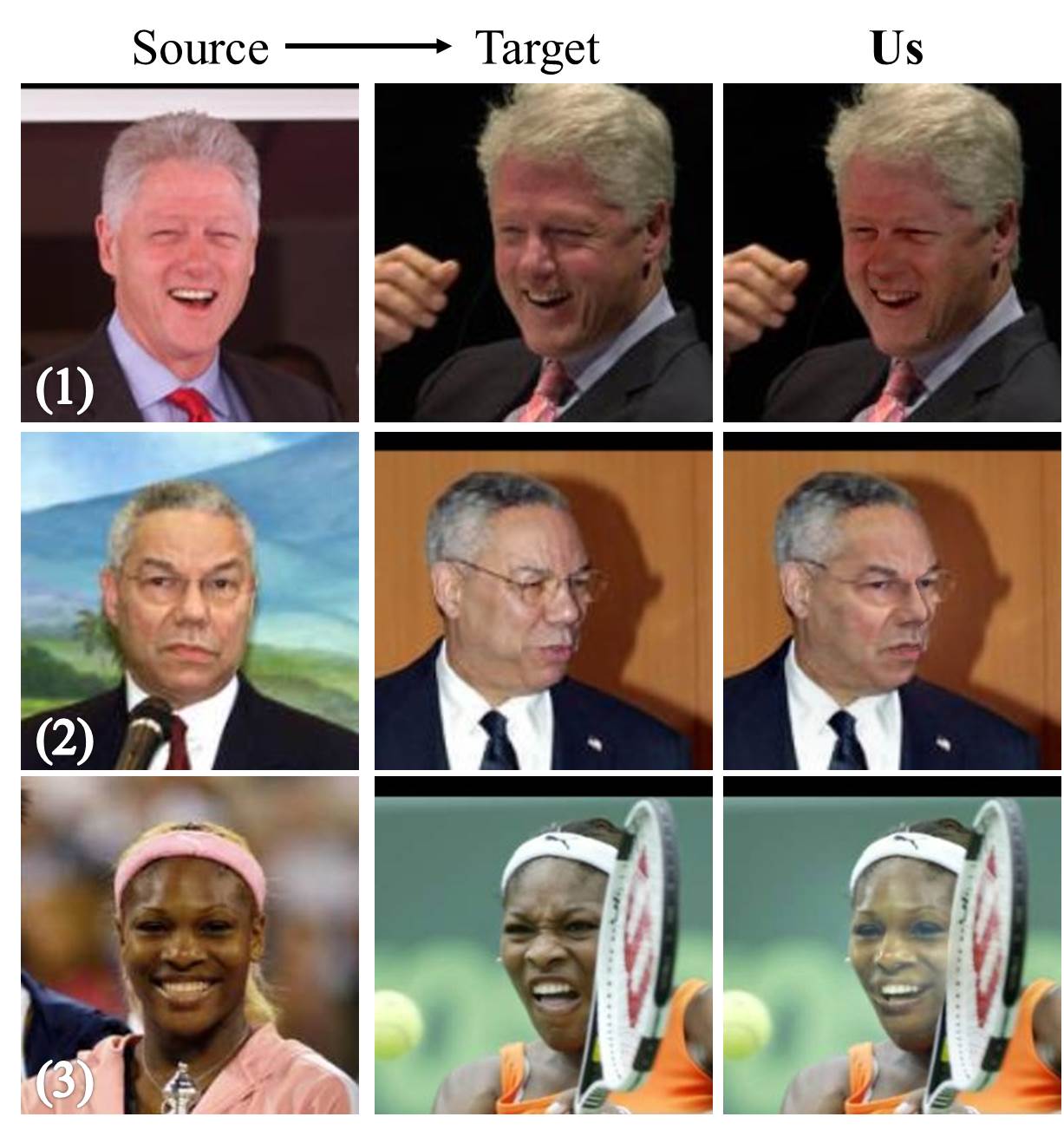}\vspace{-2mm}
\caption{
{\em Qualitative LFW intra-subject face swapping.} Faces remain recognizable despite substantially different view settings.\vspace{-4mm}}\label{fig:qual_intra}
\end{figure}

We again run our entire evaluation twice: once, swapping the first image in each test pairs keeping the second unchanged, and vice versa. Our results average these two trials. These results, obtained using different components of our system, are provided in Tab.~\ref{tab:intra} and Fig.~\ref{fig:rocs}. Example intra-subject face swap results are provided in Fig.~\ref{fig:qual_intra}. These show that even under extremely different viewing conditions, perceived subject identity remains unchanged, supporting our {\em Assumption~2}.

In general accuracy drops by $~$1\%, with a similar nAUC, compared to the use of the original, unchanged LFW images. This very slight drop suggests that our swapping does not obscure apparent face identities, when performed between different images of the same subject. Moreover, in such cases, accuracy for estimated vs. generic 3D shapes is nearly identical.

\begin{table}[t]
\centering
\resizebox{0.98\linewidth}{!}{
\begin{tabular}{l ccc}
\toprule
Method & 100\%-EER  &	Acc. &	nAUC  \\
\cmidrule(r){1-1} \cmidrule(l){2-4}
Baseline & \emph{98.10$\pm$0.90} & \emph{98.12$\pm$0.80} & \emph{99.71$\pm$0.24}\\
\cmidrule(r){1-1}
\multicolumn{4}{c}{Context preserving (face swapped out)} \\
Generic & 64.58 $\pm$ 2.10 & 64.56 $\pm$ 2.22 & 69.94 $\pm$ 2.24 \\
Est. 3D & 69.00 $\pm$ 1.43 & 68.93 $\pm$ 1.19 & 75.58 $\pm$ 2.20 \\
Seg. & 68.93 $\pm$ 1.98 & 69.00 $\pm$ 1.93 & 76.06 $\pm$ 2.15 \\
Est. 3D+Seg. & {73.17 $\pm$ 1.59} & {72.94 $\pm$ 1.39} & {80.77 $\pm$ 2.22} \\
\cmidrule(r){1-4}
\multicolumn{4}{c}{Face preserving (face swapped in)} \\
Generic & 92.28$\pm$1.37 & 92.25$\pm$1.45 & 97.55$\pm$0.71\\
Est. 3D& 88.77$\pm$1.50 & 88.53$\pm$1.25 & 95.53$\pm$0.99\\
Seg.& 89.92$\pm$1.48 & 89.98$\pm$1.36 & 96.17$\pm$0.93\\
Est. 3D+Seg.& {86.48$\pm$1.74} & {86.38$\pm$1.50} & {93.71$\pm$1.42}\\
\bottomrule
\end{tabular}
}\vspace{-2mm}
\caption{{\em Inter-subject face swapping.} Ablation study.}
\vspace{-2mm}
\label{tab:inter}
\end{table}

\begin{table}[t]
\centering
\resizebox{0.98\linewidth}{!}{
\begin{tabular}{l ccc}
\toprule
Method & 100\%-EER  &	Acc. &	nAUC  \\
\cmidrule(r){1-1} \cmidrule(l){2-4}
Baseline & \emph{98.10$\pm$0.90} & \emph{98.12$\pm$0.80} & \emph{99.71$\pm$0.24}\\
\cmidrule(r){1-1}
Generic & 97.02$\pm$0.98 & 97.02$\pm$0.97 & {99.53$\pm$0.31}\\
Est. 3D & 97.05$\pm$0.98 & 97.03$\pm$1.01 & 99.52$\pm$0.32\\
Seg. & 97.12$\pm$1.09 & 97.08$\pm$1.07 & {99.53$\pm$0.31}\\
Est. 3D+Seg.& {97.12$\pm$1.09} & {97.12$\pm$0.99} & 99.52$\pm$0.31\\
\bottomrule
\end{tabular}
}\vspace{-2mm}
\caption{{\em Intra-subject face swapping.} Ablation study.}
\vspace{-4mm}
\label{tab:intra}
\end{table}

\section{Conclusions}\label{sec:conclusions}
We describe a novel, simple face swapping method which is robust enough to allow for large scale, quantitative testing. From these tests, several important observations emerge. {\bf (1)} Face segmentation state of the art speed and accuracy, outperforming methods tailored for this task, can be obtained with a standard segmentation network, provided that the network is trained on rich and diverse examples. {\bf (2)} Collecting such examples is easy. {\bf (3)} Both faces and their contexts play important roles in recognition. We offer quantitative support for the two decades old claim of Sinha and Poggio~\cite{sinha1996think}. {\bf (4)} Better swapping, (e.g., to better mask facial spoofing attacks on biometric systems) leads to more facial changes and a drop in recognition. Finally, {\bf (5)}, 3D face shape estimation better blends the two faces together and so produces less recognizable source faces.

\section*{Acknowledgments}
This research is based upon work supported in part by the Office of the Director of National Intelligence (ODNI), Intelligence Advanced Research Projects Activity (IARPA), via IARPA 2014-14071600011. The views and conclusions contained herein are those of the authors and should not be interpreted as necessarily representing the official policies or endorsements, either expressed or implied, of ODNI, IARPA, or the U.S. Government.  The U.S. Government is authorized to reproduce and distribute reprints for Governmental purpose notwithstanding any copyright annotation thereon. TH was also partly funded by the Israeli Ministry of Science, Technology and Space

{\small
\bibliographystyle{ieee}

}

\onecolumn

\pagebreak

   \newpage
   \null
   \vskip .375in
   \begin{center}
      {\Large \bf Additional results \par}
   \end{center}
\appendix

\section{\vspace{-2mm}Additional intra-subject qualitative results}
\vspace{-2mm}
\begin{figure*}[h!]
\centering
\includegraphics[width=0.855\linewidth]{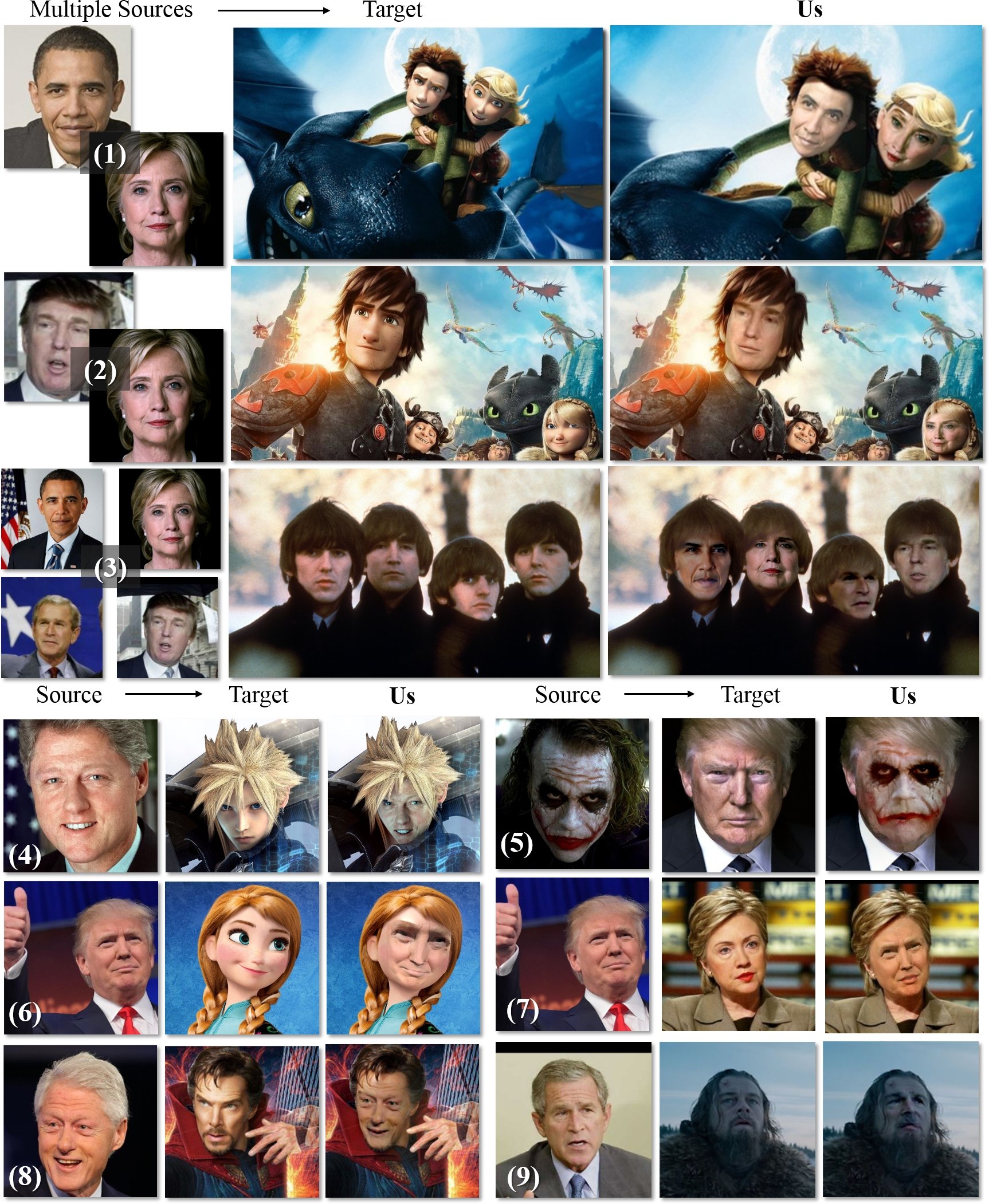}
\vspace{-2mm}
\caption{
{\em Qualitative intra-subject face swapping results.}\vspace{-5mm}}
\end{figure*}

\section{Additional segmentation results}
\begin{figure*}[h!]
\vspace{-2mm}
\centering
\includegraphics[width=0.9\linewidth]{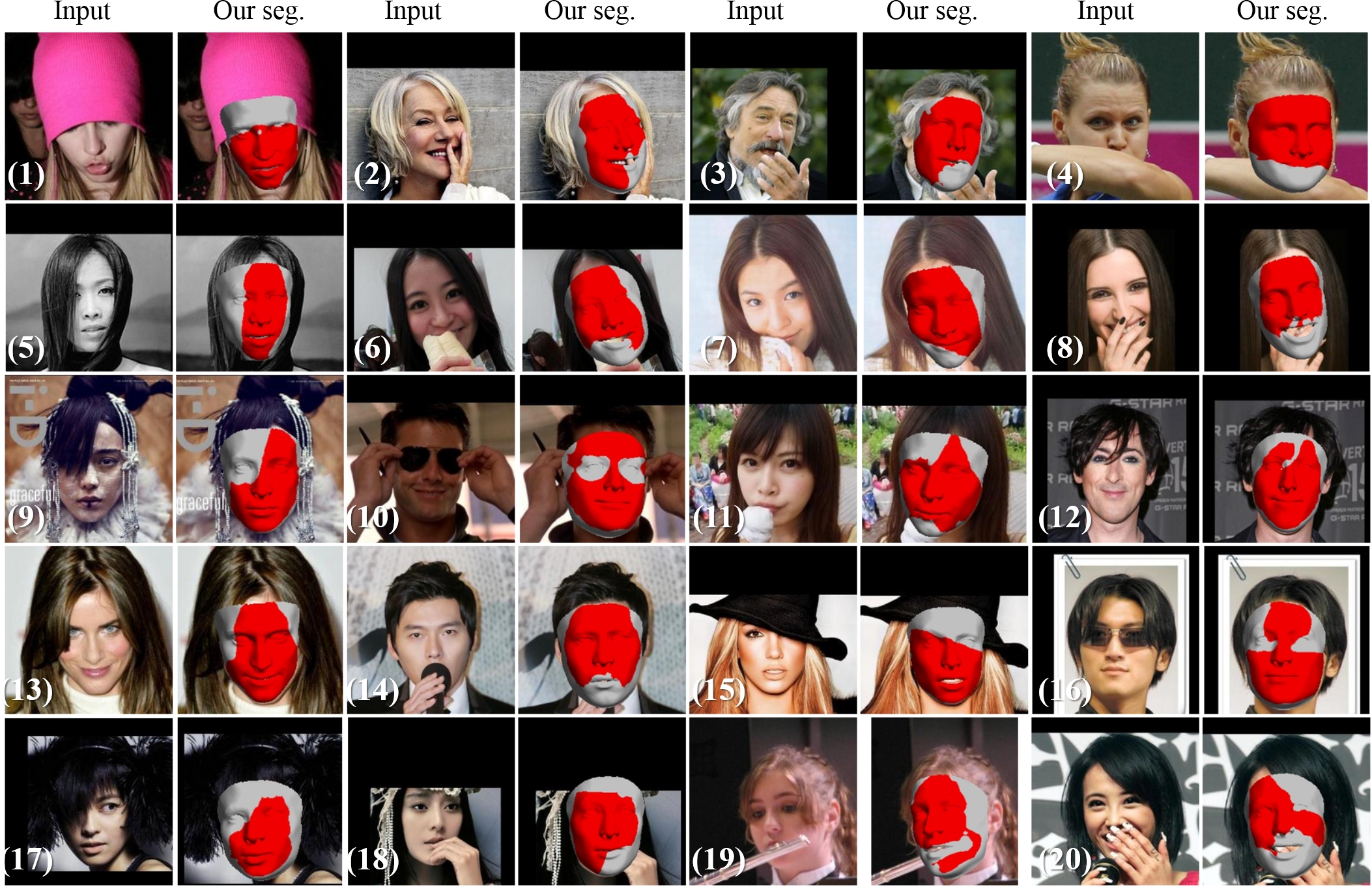}
\vspace{-2mm}
\caption{
{\em Qualitative segmentation results from the COFW data set}\vspace{-6mm}}
\end{figure*}
\begin{figure*}[h!]
\centering
\includegraphics[width=0.9\linewidth]{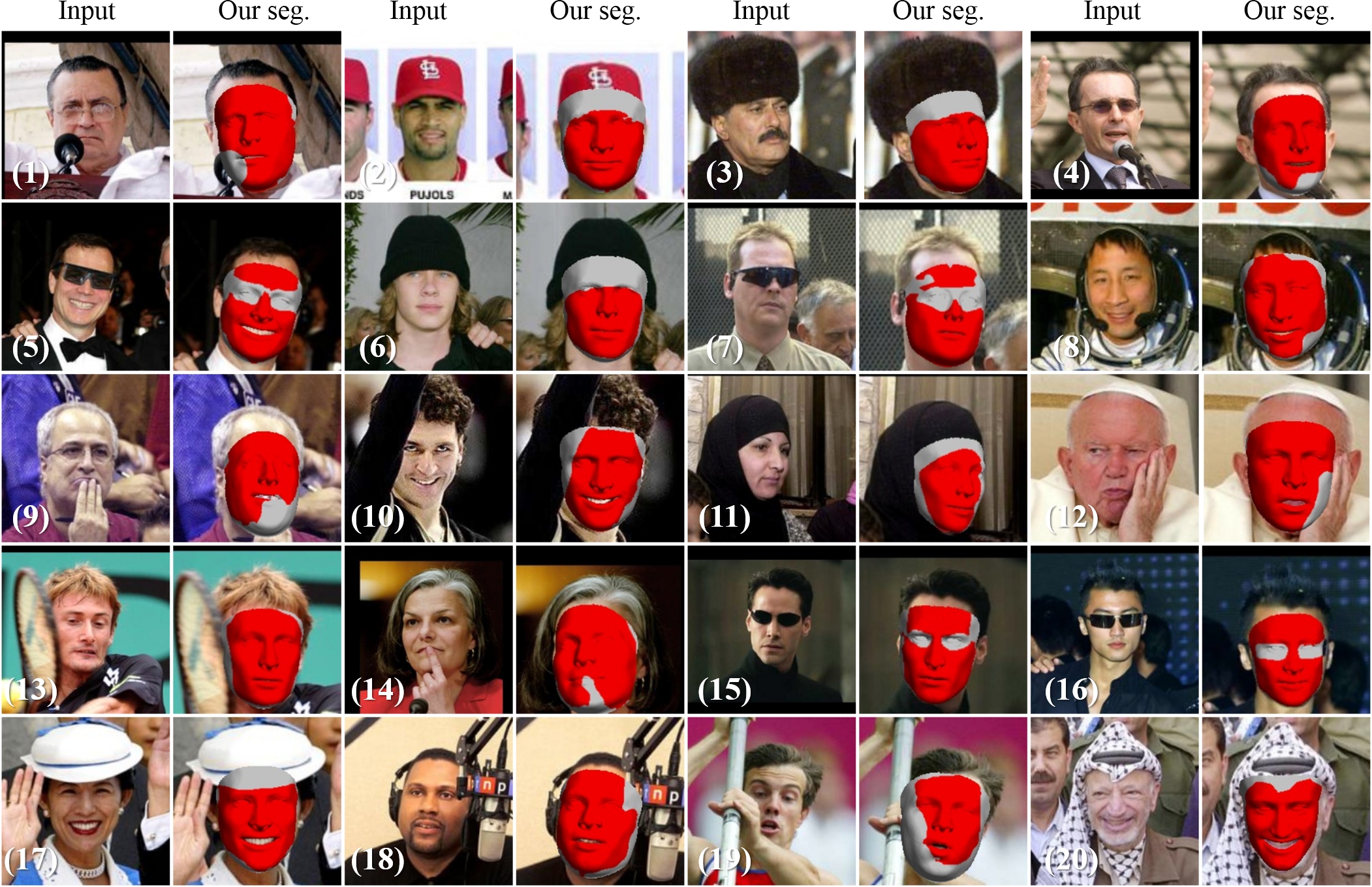}
\vspace{-2mm}
\caption{
{\em Qualitative segmentation results from the LFW data set}\vspace{-5mm}}
\end{figure*}

\section{Qualitative ablation results}
\begin{figure*}[h!]
\centering
\includegraphics[width=\linewidth]{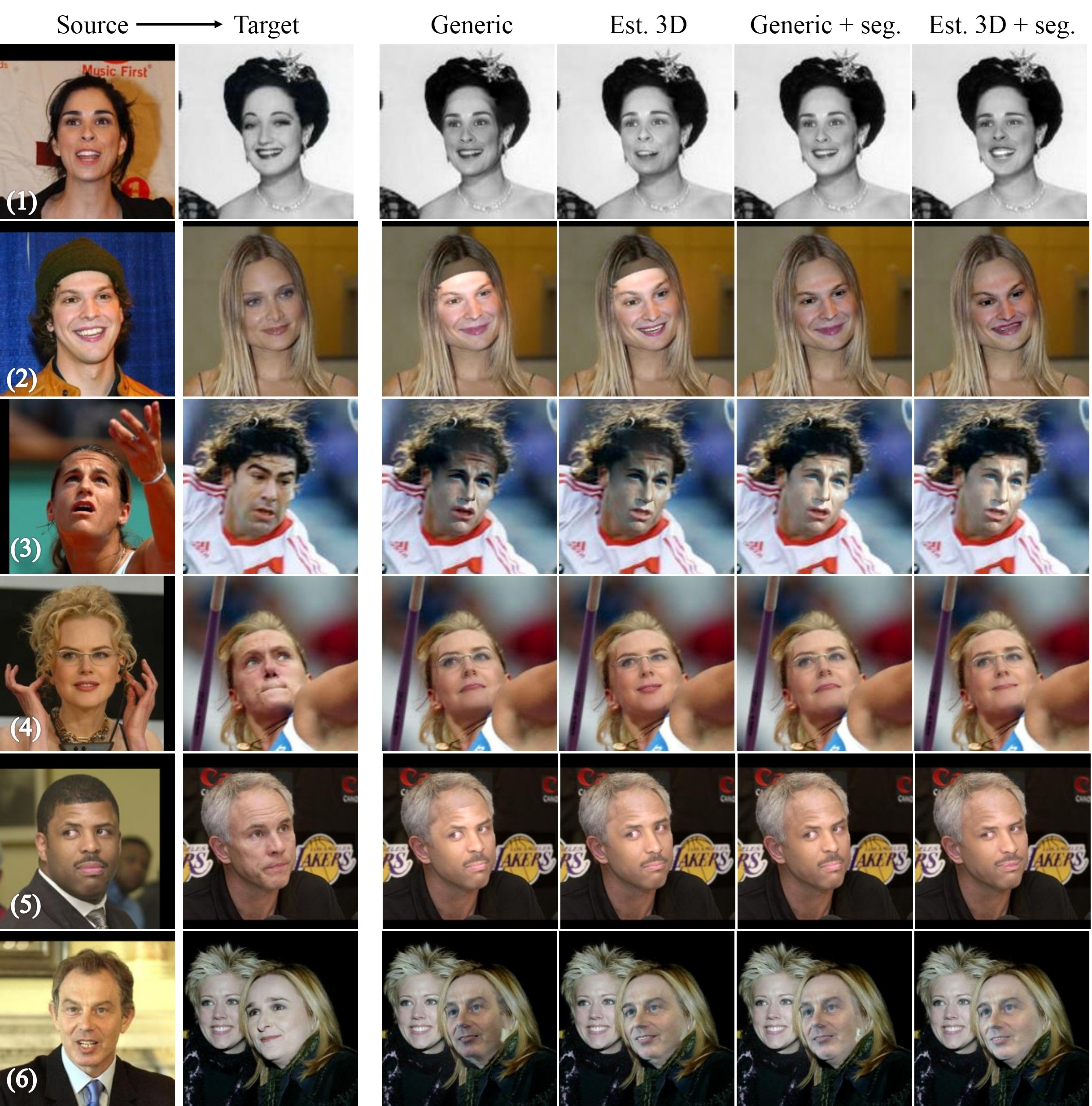} 
\vspace{-2mm}
\caption{
{\em Inter-subject face Swapping results.} Qualitative ablation study.\vspace{-5mm}}
\end{figure*}
\begin{figure*}[h!]
\centering
\includegraphics[width=\linewidth]{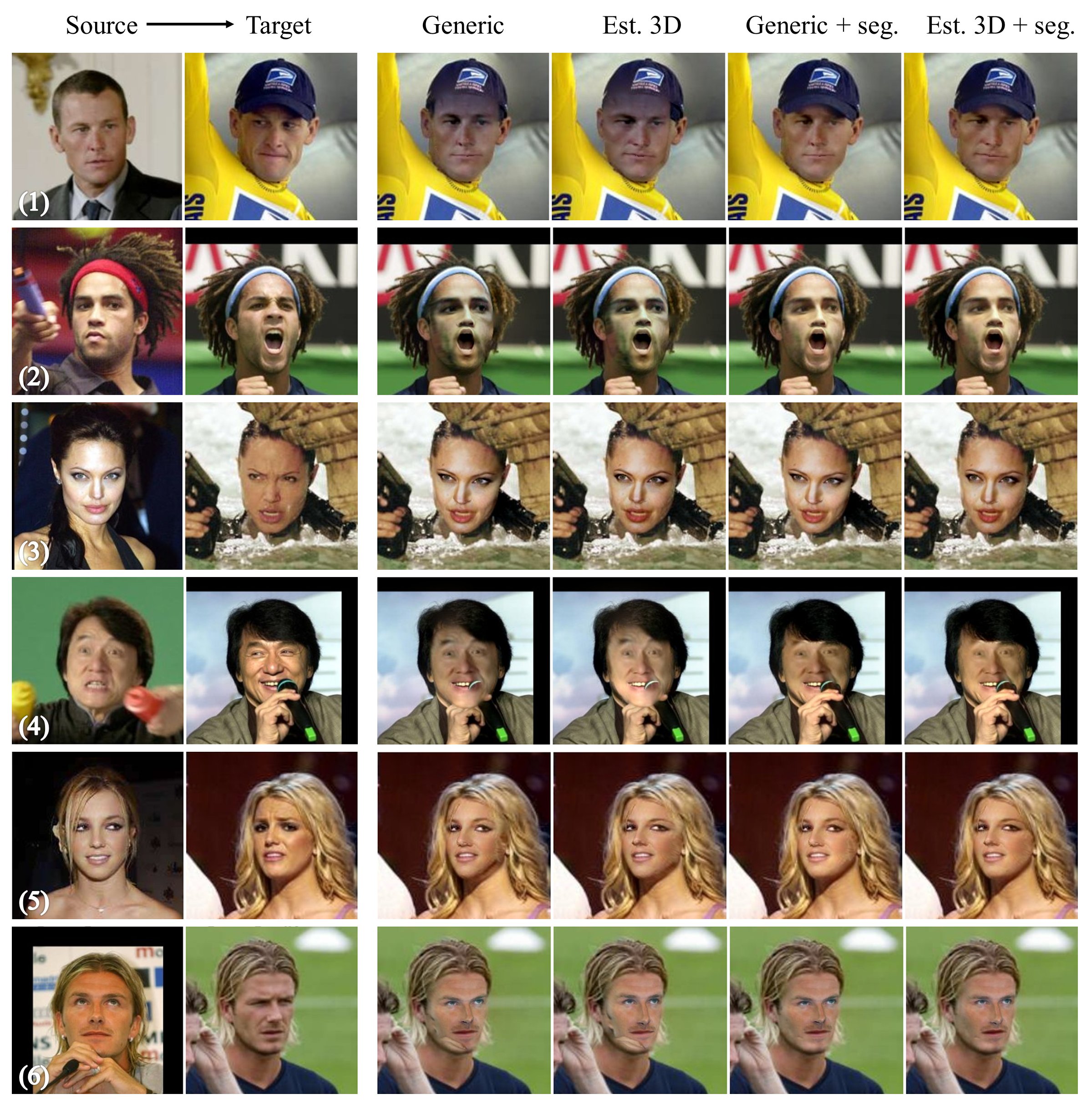}
\vspace{-2mm}
\caption{
{\em Intra-subject face swapping results.} Qualitative ablation study.\vspace{-5mm}}
\end{figure*}

\end{document}